\documentclass[10pt,twocolumn,letterpaper]{article}

\usepackage[pagenumbers]{wacv} 
\usepackage{multirow}

\definecolor{wacvblue}{rgb}{0.21,0.49,0.74}
\usepackage[pagebackref,breaklinks,colorlinks,allcolors=wacvblue]{hyperref}

\def\wacvPaperID{3170} 
\def\confName{WACV}
\def\confYear{2026}

\title{SGW-GAN: Sliced Gromov-Wasserstein Guided GANs for Retinal Fundus Image Enhancement}

\author{Yujian Xiong$^{1}$, Xuanzhao Dong$^{1}$, Wenhui Zhu$^{1}$, Xin Li$^{1}$, Oana Dumitrascu$^{2}$, Yalin Wang$^{1}$ \\
$^{1}$ Arizona State University, AZ, USA \\
$^{2}$ Mayo Clinic, AZ, USA \\
}

\begin{document}
\maketitle
\begin{abstract}

    Retinal fundus photography is indispensable for ophthalmic screening and diagnosis, yet image quality is often degraded by noise, artifacts, and uneven illumination. Recent GAN- and diffusion-based enhancement methods improve perceptual quality by aligning degraded images with high-quality distributions, but our analysis shows that this focus can distort intra-class geometry: clinically related samples become dispersed, disease-class boundaries blur, and downstream tasks such as grading or lesion detection are harmed. The Gromov Wasserstein (GW) discrepancy offers a principled solution by aligning distributions through internal pairwise distances, naturally preserving intra-class structure, but its high computational cost restricts practical use. To overcome this, we propose \textbf{SGW-GAN}, the first framework to incorporate Sliced GW (SGW) into retinal image enhancement. SGW approximates GW via random projections, retaining relational fidelity while greatly reducing cost. Experiments on public datasets show that SGW-GAN produces visually compelling enhancements, achieves superior diabetic retinopathy grading, and reports the lowest GW discrepancy across disease labels, demonstrating both efficiency and clinical fidelity for unpaired medical image enhancement.
    
\end{abstract}


\section{Introduction}

Color fundus photography (CFP) is the most widely used non-invasive retinal imaging modality, enabling large-scale screening and diagnosis of conditions such as diabetic retinopathy (DR), glaucoma, and age-related macular degeneration. Its popularity stems from convenience and low acquisition cost~\cite{dumitrascu2022automated,qian2024competition,zhu2023beyond,zhu2023self,nath2012techniques}. However, CFP images are frequently degraded by blur, uneven illumination, media opacities, color shifts, or camera-related artifacts, which hinder both expert diagnosis and automated downstream analysis including vessel segmentation, lesion detection, and disease grading~\cite{fu2019evaluation,shen2020modeling}. Improving the quality of low-quality CFP images is therefore of both clinical and algorithmic importance.

Over the past decade, deep learning, particularly generative models, has significantly advanced fundus image enhancement. Supervised methods such as SCR-Net~\cite{li2022structure}, Cofe-Net~\cite{shen2020modeling}, PCE-Net~\cite{liu2022degradation}, and GFE-Net~\cite{li2023generic} have leveraged paired low/high-quality images to learn pixel-to-pixel mappings, while RFormer~\cite{deng2022rformer} explored transformer-based architectures for better modeling of long-range dependencies. These paired methods, though powerful, are limited by the scarcity of aligned datasets, since real-world clinical acquisition rarely produces perfectly matched pairs.

As a result, unpaired enhancing have become increasingly prominent. Generative Adversarial Networks (GAN) based models~\cite{creswell2018generative,goodfellow2020generative} such as CycleGAN~\cite{zhu2017unpaired}, Wasserstein GAN (WGAN)~\cite{gulrajani2017improved}, and Optimal Transport (OT) guided GANs~\cite{zhu2023otre,zhu2023optimal} have been used to translate low-quality fundus images to a clean domain, with variants like OTEGAN~\cite{zhu2023otre} incorporating perceptual structural similarity (MS-SSIM) and identity losses, and topology-aware optimal transport (TPOT)~\cite{dong2025tpot} emphasizing preservation of vascular connectivity. Meanwhile, Stochastic Differential Equation (SDE) based approaches~\cite{song2019generative,ho2020denoising,song2020denoising} have emerged, such as CUNSB-RFIE~\cite{dong2025cunsb}, which formulates enhancement as a Schrödinger Bridge problem to ensure smooth, probabilistic transformations between domains. While these unpaired approaches alleviate the need for paired data and show promising results, recent benchmarks~\cite{zhu2025eyebench} highlight some shortcomings: conventional metrics do not adequately reflect clinical relevance, and many models struggle to consistently preserve characteristics related to downstream clinical application.

Beyond perceptual fidelity and task-level metrics, a critical yet often overlooked property of enhancement models is the preservation of \emph{intra-space geometry}. Images belonging to the same disease class should remain close in feature space after enhancement, thereby maintaining clinically meaningful relational structures. Conventional OT formulations mainly encourage alignment toward the global distribution of high-quality images, whereas the GW distance~\cite{memoli2011gromov} aligns distributions by comparing internal pairwise distances, offering a principled way to preserve intra-class relationships. GW-GAN~\cite{bunne2019learning} has illustrated the potential of this approach in generic generative modeling. However, its application to high-resolution medical CFP remains largely underexplored.

Moreover, the high computational burden of GW makes it difficult to scale in practice. To overcome this, the \emph{Sliced Gromov–Wasserstein} (SGW) distance~\cite{titouan2019sliced} provides an efficient approximation by projecting features into multiple one-dimensional subspaces and averaging the results. This retains the relational preservation properties of GW while drastically reducing computational cost, thereby making the use of GW-like constraints feasible in high-resolution tasks such as retinal fundus image enhancement.

In this work, our contributions are threefold:
\begin{itemize}
    \item We propose SGW-GAN, a novel framework that integrates SGW discrepancy into GAN training to jointly improve image quality and preserve intra-class geometry.
    \item To the best of our knowledge, we are the first to explore the use of GW/SGW in high-resolution retinal enhancement, demonstrating its ability to maintain disease-related label structures essential for clinical diagnosis.
    \item We conduct extensive experiments on public retinal datasets, showing that SGW-GAN produces visually compelling results, achieves superior diabetic retinopathy grading, and yields the lowest GW discrepancy.
\end{itemize}

\section{Related Work}



\subsection{Paired Enhancement}
Paired fundus image enhancement methods rely on supervised training with aligned low/high-quality image pairs. Representative CNN or VAE based approaches include SCR-Net~\cite{li2022structure}, Cofe-Net~\cite{shen2020modeling}, PCE-Net~\cite{liu2022degradation}, and GFE-Net~\cite{li2023generic}, which incorporate structural priors such as frequency information, Laplacian pyramids, or artifact maps to improve fidelity. Transformer-based RFormer~\cite{deng2022rformer} extends this paradigm by modeling long-range spatial dependencies. I-SECRET~\cite{cheng2021secret} further integrates semi-supervised learning, combining paired pixel-level supervision with adversarial unpaired objectives. While these paired approaches often achieve strong results on synthetic noisy-clean datasets, their applicability is limited in practice due to the difficulty of obtaining large, clinically representative paired datasets. Consequently, their improvements in PSNR and SSIM may not translate to real-world clinical reliability.

\subsection{Unpaired Enhancement}

Unpaired methods avoid the need for aligned data. GAN-based models such as CycleGAN~\cite{zhu2017unpaired} and WGAN~\cite{gulrajani2017improved} opened this direction, with extensions like OTTGAN~\cite{zhu2023optimal}, OTEGAN~\cite{zhu2023otre}, and TPOT~\cite{dong2025tpot} introducing perceptual, identity, or topology-aware regularization. While these methods improve perceptual quality, benchmarks~\cite{zhu2025eyebench} show they often distort lesion integrity. Diffusion-based approaches, such as CUNSB-RFIE~\cite{dong2025cunsb}, offer smooth probabilistic mappings but can attenuate high-frequency diagnostic features and remain computationally expensive. However, both GAN- and diffusion-based approaches tend to distort the intra-class data structure, and only limited work has explicitly addressed this issue; most methods primarily focus on denoising, while preserving label consistency across disease classes is essential for clinical reliability.

\section{Preliminary Study}
\begin{figure}[h]
\centering
\begin{subfigure}{0.48\linewidth}
    \centering
    \includegraphics[width=\linewidth]{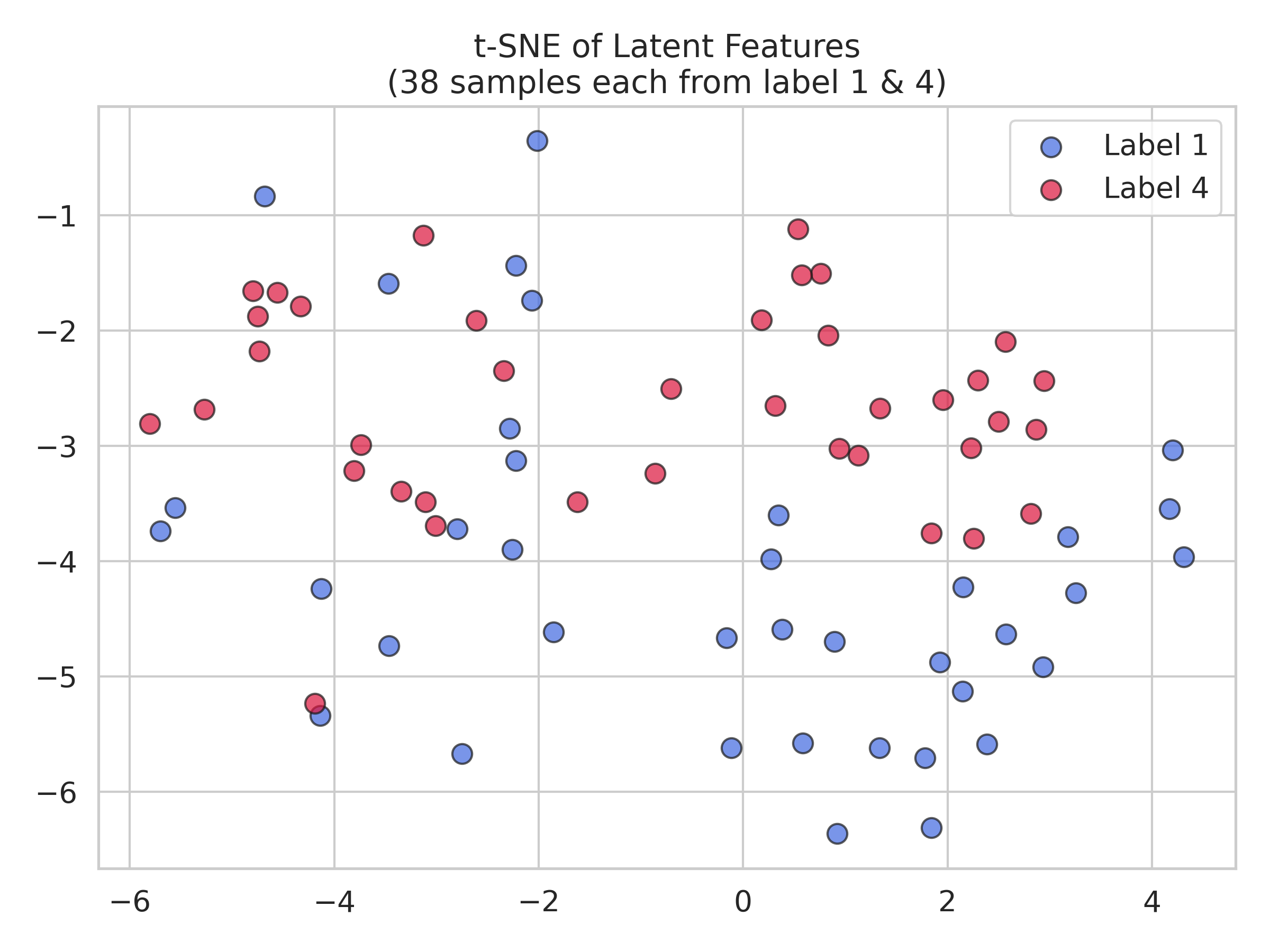}
    \caption{Low-quality images}
    \label{fig:tsne_lq}
\end{subfigure}
\hfill
\begin{subfigure}{0.48\linewidth}
    \centering
    \includegraphics[width=\linewidth]{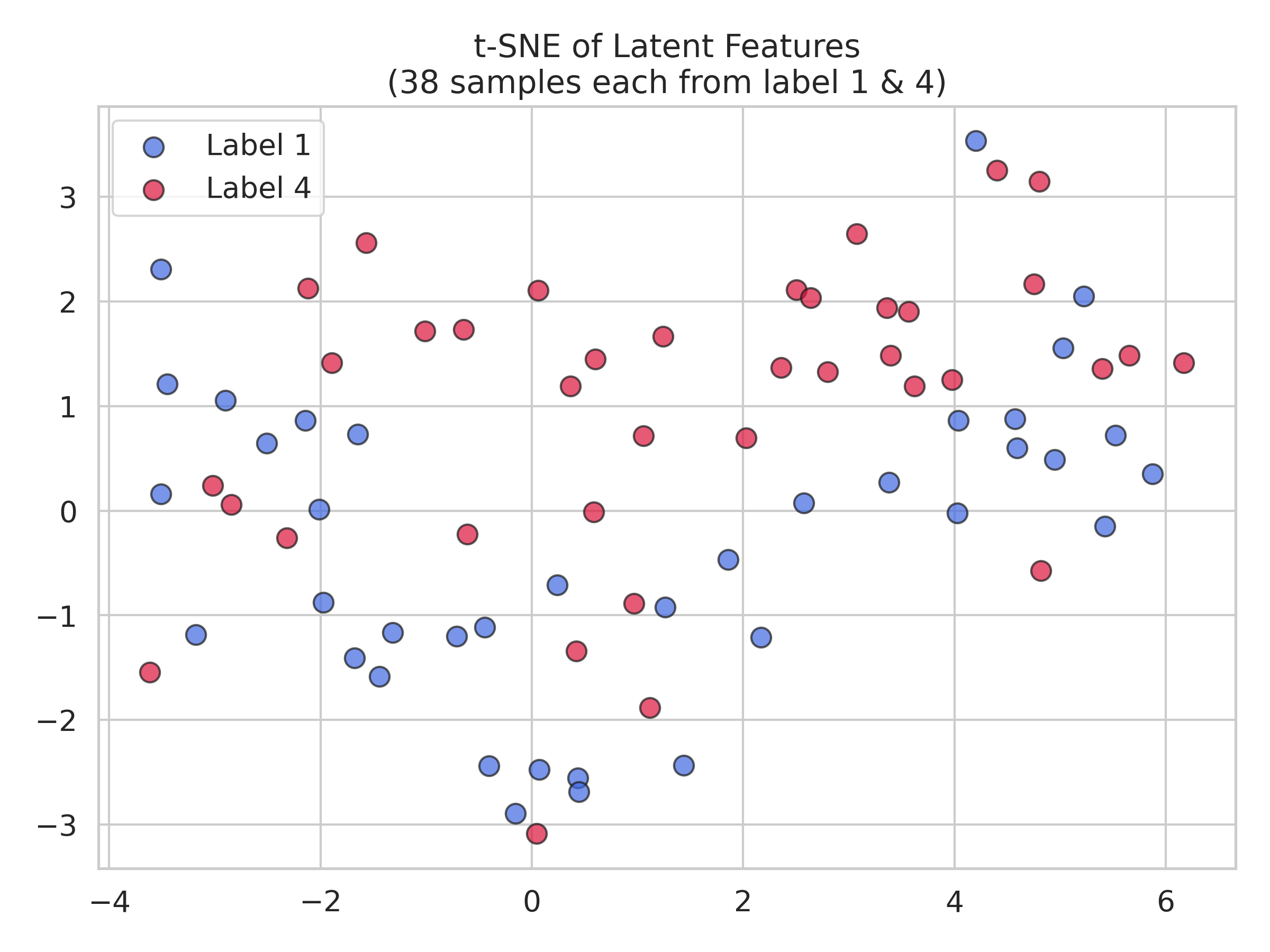}
    \caption{Enhanced fake images}
    \label{fig:tsne_fake}
\end{subfigure}
\caption{t-SNE visualizations of feature distributions across 2 disease labels, calculated from encoded feature using RetFound~\cite{zhou2023foundation}: (a) Low-quality images already show reasonable separation between labels; (b) After enhancement by OTEGAN~\cite{zhu2023otre}, the class separation become blurred, distorting the intra-space geometry.}
\label{fig:tsne_compare}
\end{figure}

Although GAN and diffusion models primarily minimize the distributional gap between low- and high-quality images, our preliminary analysis suggests that this often comes at the cost of distorting intra-class geometry. As shown in Fig.~\ref{fig:tsne_compare}, the feature distribution of low-quality images already separates disease labels reasonably well, but after enhancement by conventional methods, the class boundaries become blurred, effectively increasing inter-class confusion. This indicates that models may produce better-looking images while inadvertently harming downstream tasks such as grading or lesion detection.

\begin{figure*}[t]
\centering
\includegraphics[width=0.9\linewidth]{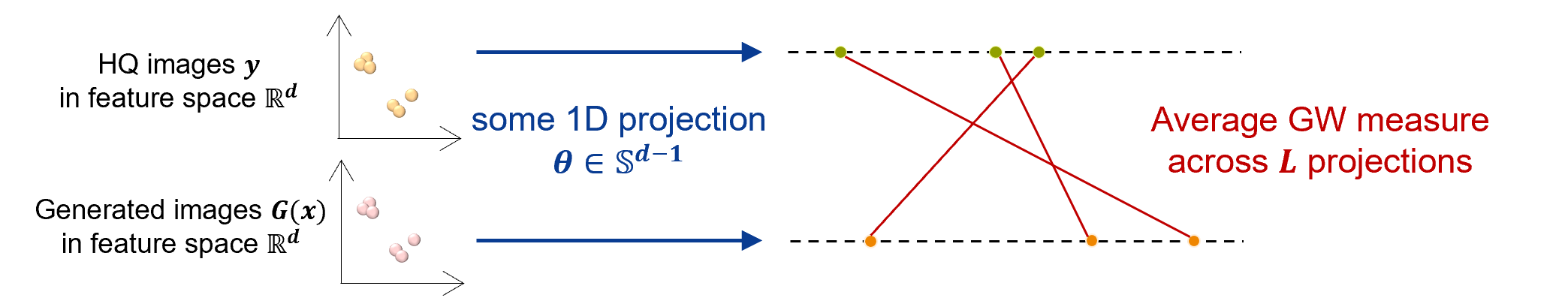}
\caption{Illustration of how SGW accelerates the GW computing process by projecting features onto multiple random one-dimensional directions, computing one-dimensional GW in each subspace, and aggregating results with multiple projections. This reduces computational cost by orders of magnitude while preserving relational fidelity.}
\label{fig:sgw}
\end{figure*}

To address this issue, we require a regulation term that explicitly enforces the intra-space geometry of enhanced images to remain close to that of the real high-quality distribution, instead of solely translate each images to a similar distribution of high-quality images. This observation naturally directs our attention to the Gromov–Wasserstein distance, which was designed to compare distributions by aligning their internal relational structures rather than only minimizing global distributional gaps.

\subsection{Gromov-Wasserstein and Sliced GW}

Unlike traditional OT metrics, which have been widely integrated into generative modeling (WGAN~\cite{gulrajani2017improved} pioneered this by replacing Jensen–Shannon divergence with the Earth Mover’s distance to improve stability), the Gromov–Wasserstein distance~\cite{memoli2011gromov} generalizes OT to incomparable metric spaces by aligning internal pairwise distances. GW-GAN~\cite{bunne2019learning} leveraged this property to regularize generative models with relational alignment, demonstrating the ability to preserve manifold structure even across unpaired and heterogeneous datasets. However, GW computations scale poorly, often cubic in sample size, limiting their use in practice.  

The Sliced GW distance~\cite{titouan2019sliced} alleviates this computational bottleneck by projecting distributions into random one-dimensional subspaces, computing GW distances efficiently in each, and aggregating the results. This reduces computational overhead by orders of magnitude while retaining relational fidelity, making SGW a practical approximation of GW for large-scale neural network training.  

Despite these advantages, applications of GW and SGW in high-resolution medical imagery remain underexplored. To the best of our knowledge, \textbf{SGW-GAN} is the first framework to demonstrate their effectiveness in retinal fundus enhancement, achieving both improved perceptual quality and preservation of intra-class label structure—an aspect that is particularly critical for clinical diagnosis.

\section{Method}

Our approach, SGW-GAN, combines the relational preservation ability of the GW discrepancy with the computational scalability of the SGW approximation, and integrates both into a GAN framework for retinal fundus image enhancement. In this section, we first review the GW discrepancy, then introduce the SGW formulation, and finally describe our adversarial strategy.

\subsection{Gromov-Wasserstein Discrepancy}

OT provides a way to compare probability distributions by finding a transport plan that minimizes a cost function between their supports. When the two distributions lie in comparable metric spaces, classical Wasserstein distance is appropriate. However, in the unpaired retinal enhancement setting, input and target distributions often lack direct correspondence: degraded and high-quality images may not be aligned one-to-one, yet they share internal relational structures (e.g., relative similarity within the same disease stage).

Let $(X,d_X,\mu_X)$ and $(Y,d_Y,\mu_Y)$ be two metric measure spaces, where $X=\{x_i\}_{i=1}^n$ and $Y=\{y_j\}_{j=1}^m$ are sample sets, $d_X,d_Y$ denote intra-domain distance functions, and $\mu_X,\mu_Y$ are probability measures supported on $X$ and $Y$. A coupling $\pi \in \Pi(\mu_X,\mu_Y)$ is a joint distribution with marginals $\mu_X$ and $\mu_Y$. The GW distance addresses this case by comparing two metric measure spaces through their intra-space relations. The squared GW discrepancy~\cite{memoli2011gromov,bunne2019learning} is formulated as:
\begin{equation}
\begin{gathered}
\mathrm{GW}^2(\mu_X,\mu_Y) = 
\min_{\pi \in \Pi(\mu_X,\mu_Y)}
\int_{X^2 \times Y^2}  \\
\big| d_X(x,x') - d_Y(y,y') \big|^2 \,
d\pi(x,y)\, d\pi(x',y'),
\label{eq:gw}
\end{gathered}
\end{equation}

Equation~\ref{eq:gw} enforces the transport plan $\pi$ to preserve relative distances within each domain: if two low-quality images are close under $d_X$, their enhanced counterparts should also be close under $d_Y$. This property makes GW particularly appealing for medical imaging tasks where relational structure among disease classes carries diagnostic value.  

From a computational standpoint, Eq.~\ref{eq:gw} involves a quartic term in the variables, since pairwise distances across all sample pairs are compared. Even with entropic regularization or Sinkhorn iterations, the complexity grows rapidly with dataset size, limiting the practical use of full GW in high-resolution imaging tasks.  

\subsection{Sliced Gromov-Wasserstein Discrepancy}

\begin{figure*}[t]
\centering
\includegraphics[width=0.9\linewidth]{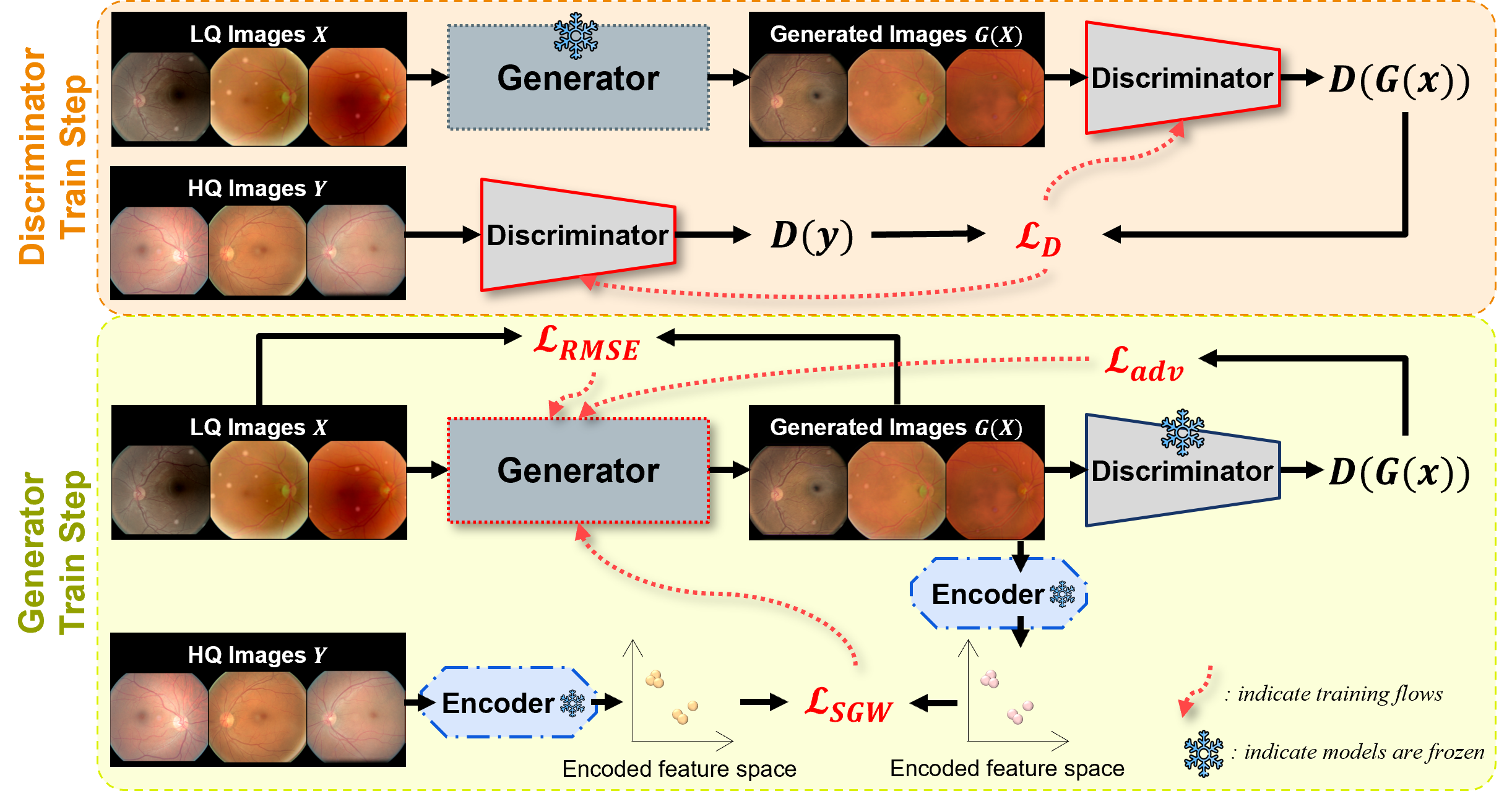}
\caption{Overview of the SGW-GAN framework. The generator produces enhanced images from low-quality inputs, while the discriminator distinguishes enhanced outputs from real high-quality images under a WGAN objective. The generator is optimized with three complementary losses: (1) RMSE regularization to preserve the eyeball structure and prevent over-enhancement, (2) SGW discrepancy to maintain intra-class relational geometry, and (3) adversarial loss to improve realism. Training alternates between updating the discriminator and the generator, ensuring that enhanced images achieve both visual quality and structural alignment with the high-quality distribution.}
\label{fig:frame}
\end{figure*}

To alleviate the heavy computational burden, recent work~\cite{titouan2019sliced} has proposed the SGW discrepancy. The core idea is to project data onto one-dimensional (1D) subspaces, compute one-dimensional GW discrepancies, and then integrate over multiple random directions. This reduces the dimensionality of the pairwise comparison problem, leading to dramatic computational savings.

Let $\theta \in \mathbb{S}^{d-1}$ be a random unit direction, and define projections $\mathcal{P}_\theta(x) = \langle x, \theta \rangle$ and $\mathcal{P}_\theta(y) = \langle y, \theta \rangle$. The projected intra-domain distances are:
\begin{equation}
\begin{aligned}
d_X^\theta(x,x') =& \big| \langle x-x', \theta \rangle \big| \\
d_Y^\theta(y,y') =& \big| \langle y-y', \theta \rangle \big|
\end{aligned}
\end{equation}
The 1D GW discrepancy under direction $\theta$ is then:
\begin{equation}
\begin{gathered}
\mathrm{GW}^2_\theta(\mu_X,\mu_Y) =
\min_{\pi \in \Pi(\mu_X,\mu_Y)} \\
\int \big| d_X^\theta(x,x') - d_Y^\theta(y,y') \big|^2 \,
d\pi(x,y)\, d\pi(x',y')
\end{gathered}
\end{equation}

Averaging across $L$, independent projections gives the SGW discrepancy:
\begin{equation}
\begin{aligned}
\mathrm{SGW}^2(\mu_X,\mu_Y) =& 
\frac{1}{L}\sum_{\ell=1}^L 
\mathrm{GW}^2_{\theta_\ell}(\mu_X,\mu_Y), \\
\quad \theta_\ell \sim& \mathrm{Unif}(\mathbb{S}^{d-1}).
\label{eq:sgw}
\end{aligned}
\end{equation}

As $L \to \infty$, the SGW distance converges to an integral representation of the full GW discrepancy, ensuring theoretical consistency. In practice, only a moderate number of projections (tens to hundreds) are needed to achieve accurate estimates. This makes SGW highly attractive for large-scale imaging tasks: while full GW scales cubically with sample size, SGW reduces the complexity close to linear in $n$ per projection, and the computations can be parallelized.  

The SGW framework inherits key properties of GW, such as invariance under isometries and the ability to compare metric spaces without direct alignment, but offers significantly faster computation. These advantages enable the integration of GW-style relational regularization into deep neural network training, which would otherwise be infeasible.

\subsection{SGW-GAN Strategy}

\begin{figure*}[t]
\centering
\includegraphics[width=0.9\linewidth]{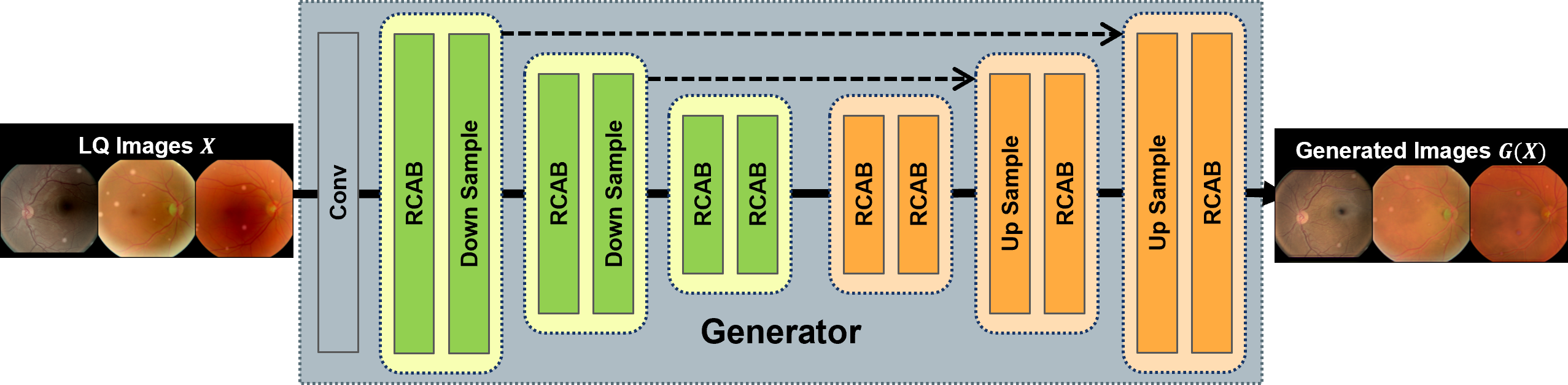}
\caption{Proposed generator structure: a U-Net with residual connection and channel attention adapted from~\cite{zhu2023optimal, wang2022optimal}.}
\label{fig:generator}
\end{figure*}

We now describe our adversarial framework, which integrates SGW into GAN training. The model is composed of a generator $G$ that maps low-quality inputs $x \sim \mu_X$ to enhanced outputs $G(x)$, and a discriminator $D$ that aims to distinguish generated images from real high-quality samples $y \sim \mu_Y$. The key idea is to jointly optimize for realism, input consistency, and relational preservation.

\subsubsection{Discriminator Training}

The discriminator follows the WGAN formulation, where the objective is to maximize the score assigned to real images while minimizing the score for generated ones. Formally, it optimizes
\begin{equation}
\mathcal{L}_D =
\mathbb{E}_{y \sim \mu_Y}[D(y)] -
\mathbb{E}_{x \sim \mu_X}[D(G(x))],
\label{eq:disc}
\end{equation}
subject to a gradient penalty that enforces the Lipschitz constraint on $D$. This regularization is critical for training stability, preventing the discriminator from saturating and ensuring meaningful gradients are propagated backwards.

\subsubsection{Generator Training}

The generator is designed to balance three complementary objectives. The first is a reconstruction term that stabilizes training by anchoring the generated image to its input. This is expressed as:
\begin{equation}
\mathcal{L}_{\mathrm{RMSE}} =
\mathbb{E}_{x \sim \mu_X} \big[ \| G(x) - x \|_2^2 \big],
\end{equation}
which penalizes large deviations from the input and prevents over-enhancement or the introduction of artificial features not present in the degraded image.

The second component is relational preservation, which incorporates the SGW discrepancy between the generated distribution $\mu_{G(X)}$ and the high-quality distribution $\mu_Y$:
\begin{equation}
\mathcal{L}_{\mathrm{SGW}} =
\mathrm{SGW}^2 \big( \mu_{G(X)}, \mu_Y \big).
\label{eq:sgw_loss}
\end{equation}
This term encourages the generator to maintain intra-space geometry, ensuring that samples corresponding to the same disease stage remain close after enhancement. Unlike pixel-level or perceptual losses, this relational loss explicitly aligns structural relationships across distributions, which is crucial for preserving disease-relevant similarity.

The third objective is adversarial realism, inherited from the GAN setting. It makes the generations indistinguishable from authentic high-quality samples by minimizing
\begin{equation}
\mathcal{L}_{\mathrm{adv}} =
- \mathbb{E}_{x \sim \mu_X}[D(G(x))].
\end{equation}
This term pushes the generator to produce outputs that fool the discriminator, thereby enforcing global visual fidelity and realism.

\paragraph{Total Objective.} The final generator objective combines these three components:
\begin{equation}
\mathcal{L}_G =
\lambda_{\mathrm{RMSE}} \mathcal{L}_{\mathrm{RMSE}} +
\lambda_{\mathrm{SGW}} \mathcal{L}_{\mathrm{SGW}} +
\lambda_{\mathrm{adv}} \mathcal{L}_{\mathrm{adv}},
\label{eq:gen}
\end{equation}
where $\lambda_{\mathrm{RMSE}}, \lambda_{\mathrm{SGW}}, \lambda_{\mathrm{adv}}$ are hyperparameters that balance input consistency, relational preservation, and adversarial realism.

Training proceeds by alternating updates: the discriminator is optimized to maximize Eq.~\ref{eq:disc}, while the generator is optimized to minimize Eq.~\ref{eq:gen}. This adversarial interaction ensures that SGW-GAN enhances visual quality, preserves intra-space geometry, and aligns generated images with the high-quality distribution. In practice, this combination stabilizes training and yields enhanced images that are both diagnostically reliable and visually convincing.

\section{Experiments and Results}

In this section we evaluate the effectiveness of SGW-GAN for retinal fundus image enhancement. We first describe the datasets and experimental setup, followed by implementation details, evaluation metrics, and comparative baselines. We then present both quantitative and qualitative results, as well as downstream task evaluations that reflect the clinical relevance of our method.

\subsection{Datasets}
We conduct experiments on the publicly available dataset EyeQ~\cite{fu2019evaluation}. All images are resized to a resolution of $256\times256$ and then center-cropped to a size of $224\times224$ before training. All input images are then normalized using mean and standard deviation according to RetFound preprocessing~\cite{zhou2023foundation,zhou2025generalistversusspecialistvision}. No other data augmentation in training the enhancement GAN model to preserve the original image patterns and structures.

\subsection{Model Details}

The generator $G$ and discriminator $D$ correspond to the implementations outlined in~\cite{zhu2023otre}. $G$ adopts a hierarchical encoder–decoder design with channel attention modules, skip connections, and a spatial attention mechanism to preserve both global structure and fine retinal details.

The generator begins with a shallow feature extraction block consisting of a $3 \times 3$ convolution followed by a channel attention block. Features are then processed through a multi-level encoder, where each stage contains stacked channel attention blocks with progressively increased feature dimensions. Downsampling between stages is implemented via bilinear interpolation followed by $1 \times 1$ convolution. The encoded features are passed into a symmetric decoder that employs skip-upsample operations, which fuse encoder features with decoder outputs via element-wise addition after attention refinement. At the output, a spatial attention module refines the reconstructed image, enforcing consistency with the input and suppressing noise artifacts. This architecture is designed to balance expressive capacity with computational efficiency.

The discriminator follows a standard convolutional design similar to PatchGAN, operating on local patches to differentiate real from generated images. It consists of stacked convolutional layers with stride $2$ for downsampling, leaky ReLU activations, and a final fully connected layer outputting a scalar response. The discriminator is trained under the WGAN formulation with gradient penalty.

For computing the SGW discrepancy, both the enhanced image $G(x)$ and the corresponding high-quality reference $y$ are passed through a frozen pretrained encoder to extract feature embeddings in $\mathbb{R}^d$ space. We employ the public \texttt{RETFound} MAE encoder~\cite{zhou2023foundation}, which has demonstrated strong retinal representation learning. The extracted embeddings are then used to compute SGW distances according to Eq.~\ref{eq:sgw_loss}. By leveraging this pretrained encoder, the SGW regularization focuses on clinically meaningful features while avoiding trivial pixel-level alignment.

\subsection{Training Details}

Training is performed with 2 Adam optimizers with learning rates of $10^{-5}$ for both discriminator and generator steps. We set $\lambda_{\mathrm{RMSE}}$, $\lambda_{\mathrm{SGW}}$, and $\lambda_{\mathrm{adv}}$ to $(100, 1000, 1)$ based on validation performance. The number of SGW projections $L$ is set to $256$. Training has 200 epochs with a batch size of 16 on a single Quadro RTX 5000 GPU for 11 hours.

\subsection{Evaluation Metrics}

We evaluate all methods using a combination of image quality measures and clinically relevant downstream tasks. To ensure fairness, all baselines are trained on EyeQ images with artificial noise and evaluated under the same protocol. The trained models are then applied to real-noise images for downstream clinical assessment.

\begin{itemize}
    \item \textbf{Image Quality:} PSNR, SSIM, and FID are computed on the EyeQ artificial-noise test set~\cite{zhu2025eyebench} to assess perceptual fidelity and distributional alignment with high-quality references. These metrics reflect general enhancement performance but are not fully indicative of clinical reliability.  

    \item \textbf{Disease Grading:} To evaluate diagnostic consistency, we trained an NN-MobileNet classifier~\cite{zhu2024nnmobilenet} for DR grading using high-quality EyeQ images. Each enhancement model is first applied to generate enhanced images from the real-noise EyeQ set, and the trained classifier is then used to infer DR severity. Performance is reported in terms of classification accuracy (ACC), F1 score, and area under the curve (AUC). This protocol primarily measures whether enhancement alters lesion distribution or disrupts disease-related features, potentially causing inconsistencies with the original grading labels.  

    \item \textbf{Relational Preservation:} Finally, we compute the full GW discrepancy between enhanced real-noise images and high-quality images. This metric directly evaluates whether intra-class geometry is preserved, ensuring that enhanced samples remain close to their disease-specific structures rather than merely improving pixel similarity.
\end{itemize}




\subsection{Baselines}
We compare SGW-GAN against other unpaired enhancement methods, include both general purpose image translation methods and CFP specified enhancement methods. CycleGAN~\cite{zhu2017unpaired}, WGAN~\cite{gulrajani2017improved}, OTTGAN~\cite{zhu2023optimal}, OTEGAN~\cite{zhu2023otre}, Context-Aware OT~\cite{vasa2025context} and CUNSB-RFIE~\cite{dong2025cunsb}. All baseline performance follows their own announced results and the recent benchmark~\cite{zhu2025eyebench}.

\section{Results}

\subsection{Quantitative Results}

\begin{table*}[h]
\centering
\caption{Quantitative comparison of retinal fundus enhancement methods on EyeQ dataset. Best results are in \textbf{bold}, second-best are \underline{underlined}. Metrics include perceptual quality (SSIM, PSNR), distributional alignment (FID), and downstream performance (ACC, F1, AUC). \textcolor{Blue}{\textbf{Notably, conventional metrics such as PSNR and SSIM, though standard in image processing, do not necessarily capture clinical relevance~\cite{zhu2025eyebench,zhu2023otre}}}. Results show that methods with lower GW discrepancy tend to achieve stronger classification, highlighting the importance of preserving intra-space distances for disease grading. SGW-GAN, while slightly lower on perceptual scores, achieves the best diagnostic performance and the lowest GW discrepancy, demonstrating superior preservation of disease-relevant features and robust distributional alignment.}
\label{tab:quality}
\begin{tabular}{lccc|ccc|c}
\toprule
\multirow{2}{*}{Methods} & \multicolumn{3}{c|}{Image Enhance Quality} & \multicolumn{3}{c|}{DR Grading} & \multirow{2}{*}{GW Discrepancy $\downarrow$} \\ 
\cmidrule(lr){2-4} \cmidrule(lr){5-7}
& SSIM $\uparrow$ & PSNR $\uparrow$ & FID $\downarrow$ & ACC $\uparrow$ & F1 Score $\uparrow$ & AUC $\uparrow$ & \\
\midrule
CycleGAN~\cite{zhu2017unpaired} & \underline{0.9313} & \textbf{25.076} & 23.778 & \underline{0.7588} & 0.7180 & 0.9251 & 807.78 \\
WGAN~\cite{gulrajani2017improved} & 0.9266 & 24.793 & 74.885 & 0.6446 & 0.6156 & 0.8874 & 886.38 \\
OTTGAN~\cite{zhu2023optimal} & 0.9275 & 24.065 & 51.201 & 0.7440 & 0.7037 & 0.9247 & 876.18 \\
OTEGAN~\cite{zhu2023otre} & \textbf{0.9392} & \underline{24.812} & 28.987 & 0.7539 & \underline{0.7228} & \underline{0.9326} & \underline{794.95} \\
Context-Aware OT~\cite{vasa2025context} & 0.9144 & 24.088 & 61.429 & 0.7301 & 0.6662 & 0.9112 & 859.28 \\
CUNSB-RFIE~\cite{dong2025cunsb} & 0.9121 & 24.242 & 33.047 & 0.6565 & 0.6341 & 0.8927 & 864.19 \\
\textbf{SGW-GAN} & 0.9035 & 23.914 & \textbf{18.141} & \textbf{0.7814} & \textbf{0.7749} & \textbf{0.9297} & \textbf{599.92} \\
\bottomrule
\end{tabular}
\end{table*}

Table~\ref{tab:quality} reports PSNR, SSIM, FID, and downstream diagnostic metrics on the EyeQ test set. Overall, SGW-GAN demonstrates a clear trade-off between perceptual quality and clinical fidelity.  

In terms of image quality, SGW-GAN achieves an SSIM of $0.9035$ and PSNR of $23.91$, slightly below CycleGAN ($0.9313$, $25.08$) and OTEGAN ($0.9392$, $24.81$). However, it significantly improves distributional alignment, reaching an FID of $18.14$ compared with $23.78$ for CycleGAN and $28.99$ for OTEGAN.  

For downstream DR grading, SGW-GAN delivers the best results with an ACC of $0.7814$, F1 score of $0.7749$, and AUC of $0.9297$, surpassing OTEGAN ($0.7539$, $0.7228$, $0.9326$). These improvements indicate that although perceptual metrics are modest, SGW-GAN preserves disease-relevant features more effectively, boosting diagnostic performance.  

Finally, SGW-GAN reports the lowest GW discrepancy ($599.9$) compared with OTEGAN ($794.9$), showing that enhanced images remain structurally closer to the high-quality distribution.  

Taken together, these results confirm that while SGW-GAN sacrifices a small margin in PSNR and SSIM, it achieves superior distributional alignment and diagnostic performance. This highlights the limitation of perceptual scores, which do not fully satisfy clinical requirements, and demonstrates the strength of SGW-GAN in preserving intra-class data structure critical for medical applications.

\begin{figure*}[t]
\centering
\includegraphics[width=\linewidth]{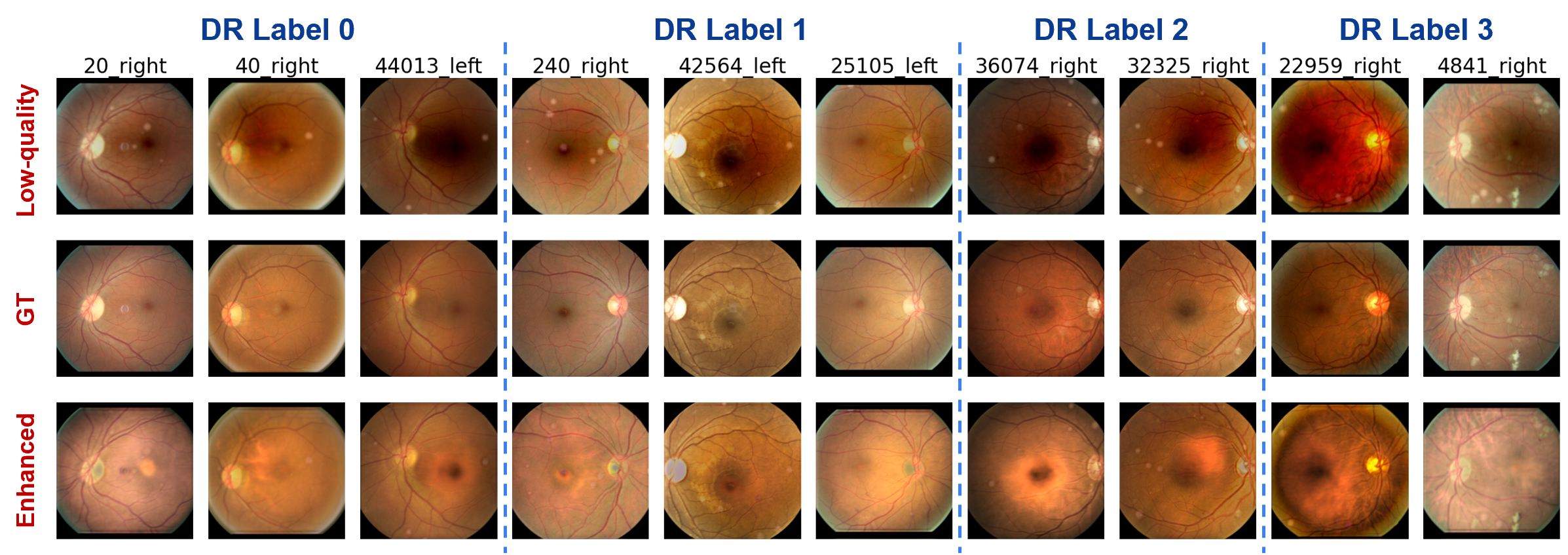}
\caption{Representative examples of retinal image enhancement. Each triplet shows the low-quality input (top row), the ground-truth high-quality image (middle row), and the enhanced output by our method (bottom row). Despite slightly lower SSIM/PSNR values compared to previous approaches, our method preserves fine vessels and overall retinal structure, while effectively removing black dots and artifacts, leading to clearer and more clinically usable images.}
\label{fig:sample}
\end{figure*}

\subsection{Enhancement Results}

Figure~\ref{fig:sample} illustrates representative examples of image enhancement obtained with our proposed method. Although the reported SSIM and PSNR values are slightly lower than those of conventional baselines, the qualitative assessment shows that our approach produces visually superior and clinically meaningful results. In particular, fine vascular structures are better preserved, the central macular region remains clear, and common degradation artifacts such as artifacts, unbalanced illumination and peripheral blotches are effectively suppressed. Compared to previous methods, our enhancements avoid over-smoothing and maintain both global retinal context and local diagnostic cues.

Another noteworthy property is that our results exhibit greater consistency across different disease categories: mild and severe DR images are enhanced in a manner that preserves their relative differences, avoiding the artificial homogenization sometimes introduced by GAN- or diffusion-based models. This aligns with our preliminary analysis and quantitative results, where conventional enhancement blurred class boundaries, whereas SGW-based regularization maintained intra-class geometry.

Overall, these characteristics confirm that numerical similarity scores alone do not fully capture the advantages of our method. The combination of structural preservation, artifact suppression, and relational consistency makes SGW-GAN particularly suitable for downstream disease grading and lesion detection tasks.

\subsection{Ablation Study}

To better understand the role of each component in our objective, we performed an ablation study by disabling one loss term at a time. The results are summarized in Table~\ref{tab:ablation}.  

\begin{table}[h]
\centering
\caption{Ablation study on the effect of each loss term. 
The full SGW-GAN achieves the best balance of perceptual quality and lowest GW discrepancy. 
Removing RMSE degrades structural fidelity of the eyeball, while removing SGW greatly increases GW discrepancy, confirming its role in preserving intra-class geometry.}
\label{tab:ablation}
\begin{tabular}{lccc}
\toprule
Model Variant & SSIM & PSNR & GW Discrepancy \\
\midrule
Full SGW-GAN & 0.9035 & 23.914 & 599.92 \\
w/o RMSE & 0.8535 & 19.914 & 621.26 \\
w/o SGW & 0.9288 & 24.417 & 847.68 \\
\bottomrule
\end{tabular}
\end{table}

When the RMSE term is removed, the enhanced images tend to exhibit structural distortions, particularly around the circular boundary of the eyeball and the global retinal layout. This shows that RMSE serves as a structural anchor, ensuring that the generator does not introduce artifacts that compromise the overall geometry of the eye.  

Eliminating the SGW loss causes a significant increase in GW discrepancy, indicating that relational consistency between the enhanced distribution and the ground truth distribution is lost. This confirms that SGW is the key factor that enforces intra-class geometry preservation and drives the final alignment with the high-quality space.

Overall, each loss contributes complementary strengths: RMSE maintains global eye structure, SGW ensures distributional alignment and intra-space consistency, and adversarial loss enhances visual realism. Removing anyone will introduce different drawback, validating the necessity of our three-term objective.

\section{Discussion}

While SGW-GAN demonstrates clear advantages in preserving disease-relevant relational structure and achieving strong downstream diagnostic performance, several limitations remain. First, the improvement in classification metrics comes at the cost of a slight reduction in perceptual quality scores (PSNR/SSIM). Although this reduction mainly reflects differences in global contrast and texture illusions rather than true diagnostic degradation, it may be perceived negatively in purely visual assessments.  

Second, the computation of SGW, although significantly more efficient than full GW, still introduces additional overhead compared to standard GAN training. In practice, this requires careful balancing of the number of projections $L$ to trade off computational cost against relational fidelity. Large-scale deployment on ultra-high-resolution images may further challenge runtime efficiency.  

Third, our framework relies on a pretrained encoder (RETFound) to extract meaningful features for SGW loss. The choice of encoder influences the relational space in which SGW is applied. Although RETFound provides strong generalization across retinal datasets, the dependency on a fixed encoder may limit adaptability to domains with very different characteristics.  

Finally, our evaluations focus on retinal fundus images. While the principles of SGW-GAN are broadly applicable, further validation on other medical and natural image domains is necessary to establish robustness and generality.

\section{Conclusion and Future Work}

In this paper, we proposed SGW-GAN, a novel framework that integrates SGW discrepancy into adversarial training for unpaired CFP image enhancement. Unlike conventional enhancement methods that focus solely on perceptual quality, SGW-GAN explicitly enforces intra-space alignment between enhanced and high-quality distributions. This design allows the model to produce enhanced images that preserve intra-space geometry, yielding superior performance on clinically relevant downstream tasks while maintaining competitive visual quality.  

The key novelty of SGW-GAN lies in the use of SGW as an efficient and scalable proxy for GW discrepancy. By leveraging random 1D projections, SGW dramatically reduces computational cost while retaining the theoretical strengths of GW. This makes it feasible to incorporate relational preservation into end-to-end neural network training. Our experiments confirm that SGW-GAN achieves the best trade-off between perceptual enhancement, structural fidelity, and diagnostic consistency.

Looking ahead, SGW-GAN has the potential to be extended beyond retinal image enhancement to any unpaired enhancement or translation problem where maintaining intra-distribution structure is critical for downstream tasks. Examples include enhancement of histopathology slides for cancer grading, denoising of MRI scans for lesion detection, and even natural image applications such as style transfer or domain adaptation, where preserving relational structure is as important as visual fidelity.  

Future work will focus on reducing computational overhead by exploring adaptive projection strategies for SGW, investigating encoder-agnostic relational spaces, and extending evaluations to broader datasets and modalities. We believe that the integration of efficient relational constraints like SGW will open new avenues for robust and clinically meaningful image enhancement in unpaired settings.


{
    \small
    \bibliographystyle{ieeenat_fullname}
    \bibliography{main}
}

\end{document}